\documentclass[10pt,twocolumn,letterpaper]{article}

\usepackage[preprint]{cvpr}      
\usepackage{booktabs}
\usepackage[table]{xcolor}
\definecolor{cvprblue}{rgb}{0.21,0.49,0.74}
\usepackage[pagebackref,breaklinks,colorlinks,allcolors=cvprblue]{hyperref}
\usepackage{algorithm}
\usepackage{algorithmic}
\usepackage[most]{tcolorbox}

\def\confName{CVPR}
\def\confYear{2026}

\newcommand{\name}{LongVideo-R1\xspace}

\title{\name: Smart Navigation for Low-cost Long Video Understanding}

\author{Jihao Qiu\textsuperscript{1}, ~Lingxi Xie\textsuperscript{2}, ~Xinyue Huo\textsuperscript{2}, ~Qi Tian\textsuperscript{2}\footnotemark[1], ~Qixiang Ye\textsuperscript{1}\footnotemark[1]\\
\textsuperscript{1}University of Chinese Academy of Sciences \quad 
\textsuperscript{2}Huawei Consumer Business Group\\
{\tt\small qiujihao19@mails.ucas.ac.cn}\quad{\tt\small 198808xc@gmail.com}\quad{\tt\small xinyueh@mail.ustc.edu.cn}\\
{\tt\small tian.qi1@huawei.com}\quad{\tt\small qxye@ucas.ac.cn}
}

\begin{document}
\twocolumn[{
\maketitle

\vspace{-25pt}
\begin{center}
\centering
\includegraphics[width=\textwidth]{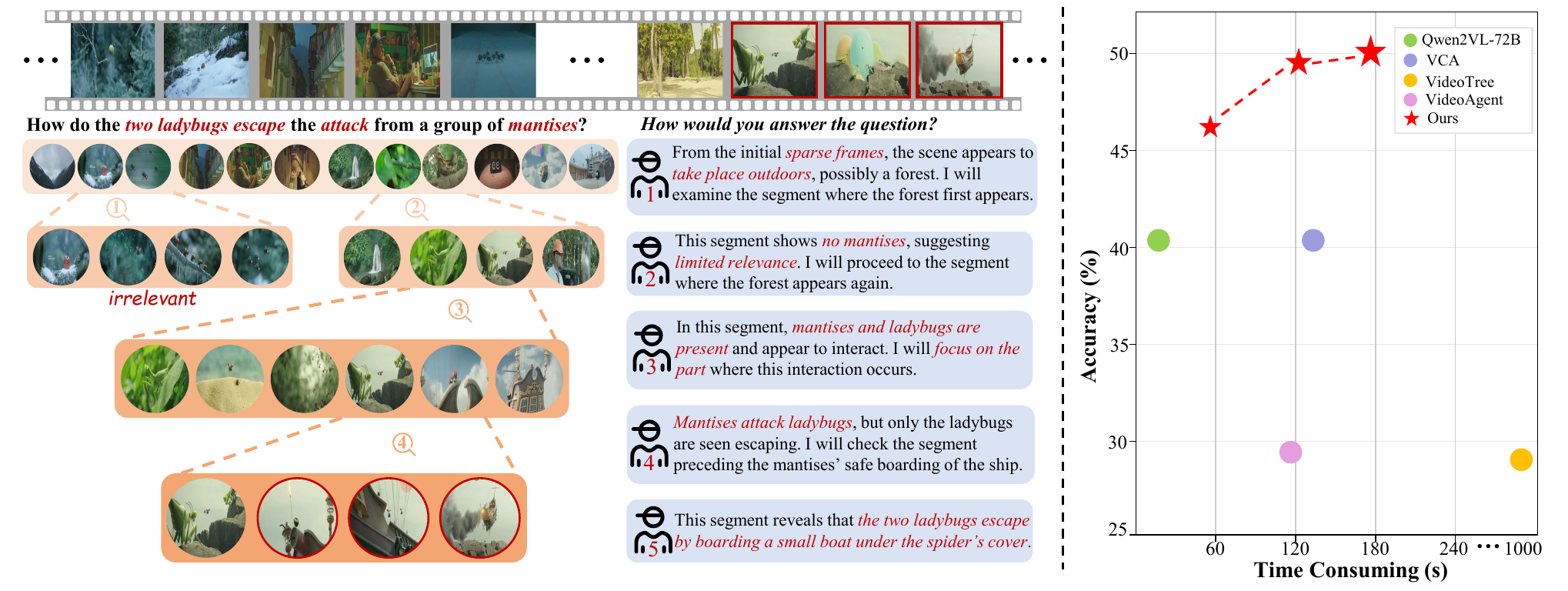}
\captionof{figure}{\textbf{Motivation and performance comparison.} 
\textbf{Left:} For efficient understanding of long video, the algorithm shall learn to fetch and perceive information \textbf{effectively}, where the core abilities are: (1) judging whether collected information is sufficient for answering, and (2) if not, navigating to the next clip most likely to contain useful information. Drawing style was inspired by Ego-R1~\cite{tian2025ego}.
\textbf{Right:} \name achieves a better tradeoff compared to recent methods on the LVBench dataset~\cite{wang2025lvbench}. The marker size indicates model scale.}
\label{fig:teaser}
\end{center}
}]

{\let\thefootnote\relax\footnotetext{* Corresponding authors.}}

\begin{abstract}
This paper addresses the critical and underexplored challenge of long video understanding \textbf{with low computational budgets}.
We propose \textbf{\name}, an active, reasoning-equipped multimodal large language model (MLLM) agent designed for efficient video context navigation, avoiding the redundancy of exhaustive search.
At the core of \name lies a reasoning module that leverages high-level visual cues to infer the most informative video clip for subsequent processing.
During inference, the agent initiates traversal from top-level visual summaries and iteratively refines its focus, immediately halting the exploration process upon acquiring sufficient knowledge to answer the query.
To facilitate training, we first extract hierarchical video captions from CGBench, a video corpus with grounding annotations, and guide GPT-5 to generate 33K high-quality chain-of-thought-with-tool trajectories. 
The \name agent is fine-tuned upon the Qwen-3-8B model through a two-stage paradigm: supervised fine-tuning (SFT) followed by reinforcement learning (RL), where RL employs a specifically designed reward function to maximize selective and efficient clip navigation.
Experiments on multiple long video benchmarks validate the effectiveness of name, which enjoys superior tradeoff between QA accuracy and efficiency. 
Code and data are available at \url{https://github.com/qiujihao19/LongVideo-R1}.
\end{abstract}

\section{Introduction}
\label{introduction}

The rapid advancement of multimodal large language models (MLLMs) has opened an unprecedented avenue for the semantic understanding of video data~\cite{maaz2024video,lin2024video}.
However, the MLLMs' success in the domain of long-form videos (those spanning 1--2 hours) is obstructed by their finite size of context, making them unable to ingest the rich visual content for comprehensive understanding.
This intrinsic limitation forces current methodologies to rely on a costly, brute-force pipeline\textemdash partitioning video to short clips, processing each clip exhaustively (\textit{e.g.}, generating captions or summarizing events), and finally integrating the results into the final answer. 
Recent studies such as Ego-R1~\cite{tian2025ego} and Videotree~\cite{wang2025videotree} reported competitive long video QA accuracy, but their complexity grows linearly with the video's length, leading to prohibitively high computational cost and latency.
This severely restricts the deployment of MLLMs in real-world applications, such as embodied agents requiring low-latency world reactions and high-throughput video-chat services constrained by per-sample processing budgets.

In this study, we introduce a new, practically motivated research setting: long video understanding under fewer computational constraints.
Instead of solely optimizing for question answering (QA) accuracy, we propose that a better measure of model efficacy lies in its ability to achieve a better accuracy-efficiency tradeoff.
We formally quantify the computational burden by accumulating the estimated cost of every operation that an MLLM requires to derive an answer. 
In other words, the objective is to find the Pareto-optimal solution where competitive accuracy is maintained with minimal computational expenditure.
The key to unlocking this efficiency is replacing exhaustive search with goal-oriented reasoning. 
We hypothesize that an MLLM must possess the ability to perform dynamic and iterative reasoning: based on partial, high-level context, it must decide which clip to sample next to locate the critical event pertaining to the question.

Motivated by the idea, we propose \textbf{\name}, a novel framework that integrates an MLLM with a large reasoning model (LRM) for smart video navigation. The long video is organized into a hierarchical structure, enabling the LRM to rapidly shift its focus across temporal granularity levels. Given a question, LongVideo-R1 begins its exploration at the top layer and, at each step, calls a video captioning tool to gather local context, and then calls a thinking module to determine whether or not the answer can be derived. If yes, a video QA tool is called to generate the final answer; otherwise, the thinking module dictates the next sampling location\textemdash it may drill down to a child clip, traverse laterally to a sibling, or backtrack to an upper layer for renewed context. The process terminates upon reaching a maximum iteration limit.

To train \name, we construct a high-quality dataset of 33K reasoning episodes leveraging the grounding annotations of the CGBench dataset~\cite{chen2024cg} and synthesize explicit reasoning trajectories using the GPT-5 API. We train the Qwen3-8B~\cite{yang2025qwen3} model using supervised fine-tuning (SFT) followed by reinforcement learning (RL) with a novel reward mechanism designed specifically to prioritize efficient navigation and accurate grounding results. The training procedure is efficient upon pre-extracted captions and stable throughout a few training epochs.

We test \name on three challenging long video QA benchmarks, \textit{i.e.}, LVBench~\cite{wang2025lvbench}, VideoMME~\cite{fu2025video}, and MLVU~\cite{zhou2025mlvu}. The results show that \name achieves competitive QA accuracy with an average of $10.5$ rounds of reasoning and navigation/answering, resulting in a significantly lower computational cost than the linear-scan methods. Furthermore, we showcase its capability for ultra-long video understanding on complex TV dramas, a domain previously inaccessible under strict budget constraints.

\section{Related Work}
\label{related_work}

\textbf{Multimodal large language models} (MLLMs)~\cite{hurst2024gpt,chen2024internvl,wang2024qwen2} represent a paradigm shift in computer vision research. Inheriting the robust reasoning capabilities of large language models (LLMs)~\cite{touvron2023llama,achiam2023gpt,team2023gemini,liu2024deepseek}, MLLMs extend this competency to the visual domain by encoding visual inputs into discrete tokens and integrating them into the model's textual context~\cite{liu2023llava,liu2024llavanext,tian2024chatterbox,qiu2024artemis} and have transcended conventional, bounded visual recognition tasks (\textit{e.g.}, classification, detection) to enable complex, open-world question answering (QA) over video data~\cite{maaz2024video,lin2024video}.

As visual understanding performance approaches saturation on static images and short video clips, the community's focus has substantially shifted toward long-form video understanding. The introduction of large-scale benchmarks featuring hour-long videos and complex QA tasks (\textit{e.g.}, EgoSchema~\cite{mangalam2023egoschema}, LongVideoBench~\cite{wu2024longvideobench}, Video-MME~\cite{fu2025video}, LVBench~\cite{wang2025lvbench}, CG-Bench~\cite{chen2024cg}, \textit{etc.}) poses significant challenges to MLLMs. Two lines of research were conducted to overcome the inherent context length limitations of MLLMs. One direction focuses on devising efficient video representations~\cite{li2024vidtome,lee2024video,shen2024tempme} to maximize the information density~\cite{shen2024longvu,choudhury2024don,zhang2024flash}. Another direction, which is highly scalable, involves segmenting the video, processing components separately, and integrating the resulting information for final inference~\cite{song2025moviechat+,tang2025adaptive,zhang2025silvr}. This latter approach has been further refined by the advent of large reasoning models (LRMs)~\cite{hurst2024gpt,guo2025deepseek,yang2025qwen3}, leading to the development of agent-based video understanding systems~\cite{fan2024videoagent,wang2024videoagent,wang2025videotree}. In these systems, an LLM agent employs explicit thinking and reasoning to strategically invoke various specialized tools, a methodology that currently dominates performance across many leading benchmarks~\cite{feng2025video,tang2025video}.

Notwithstanding the rapid progress in achieving high accuracy, relatively minimal effort has been dedicated to reducing the computational budget of long-form video understanding. For instance, recent agentic architectures, such as video-SALMONN 2~\cite{tang2025video}, and Ego-R1~\cite{tian2025ego}, necessitate the exhaustive processing of all or a substantial proportion of video segments, demanding an inordinate number of MLLM calls and consequently imposing severe computational overhead. In this paper, we formally address this deficiency by defining the pursuit of the accuracy-efficiency Pareto-optimum and subsequently introducing a competitive, agent-based solution for the pursuit.

Training a smart agent necessitates with advanced reinforcement learning techniques. Classical algorithms, notably Proximal Policy Optimization (PPO)~\cite{schulman2017proximal}, have been extended into Group Relative Policy Optimization (GRPO)~\cite{shao2024deepseekmath} to obviate the explicit reliance on a critic model, improving the efficiency of policy optimization. Numerous subsequent iterations have been proposed to refine policy learning for both LLMs~\cite{yu2025dapo,zheng2025group,zhao2025geometric} and MLLMs~\cite{huang2025vision,zhan2025vision,feng2025video,li2025videochat}. A predominant theme across these advancements is the engineering of specialized reward functions tailored to guide agent behavior toward desired outcomes.

\section{On Efficient Long Video Understanding}
\label{problem}

\subsection{What Makes Efficient Video Understanding?}
\label{problem:formulation}

Given that agentic algorithms for long-form video understanding necessitate a multi-stage process (including data preparation, clip navigation, hierarchical reasoning, and final inference), we formally define the total computational cost required for a single QA task. This cost is computed by aggregating the estimated computational overhead incurred at every step within the operational pipeline. Our primary research objective is to devise an algorithmic solution that attains a Pareto-optimal tradeoff between QA accuracy and computational efficiency\footnote{We specifically assume a setting where each QA task is executed individually and on-demand. This explicitly excludes algorithms that rely upon extensive video preprocessing, as such approaches do not satisfy the low-latency requirements of reactive or budget-constrained systems.}.

To achieve this goal, we introduce \name, a dynamic, active exploration framework. As depicted in Figure~\ref{fig:teaser}, \name operates via a self-regulating, closed-loop mechanism instantiated by two core functionalities: (1) contextual exploration, which governs the active navigation and information collection within the hierarchical video structure, and (2) reasoning and termination control, which judges the sufficiency of the gathered context for QA and, if necessary, determines the subsequent step for exploration. This iterative paradigm, where the process continues until a definitive answer is produced (or maximum iterations are reached), provides a dramatic reduction in computational expenditure compared to exhaustive search algorithms, while preserving a competitive QA accuracy.

\subsection{\name Framework}
\label{problem:framework}

The input of \name consists of a long-form video $\mathbb{V}$ and a question $\mathbf{q}$. Let us denote the duration of $\mathbb{V}$ as $T$ (in seconds); given $0\leqslant t_1<t_2\leqslant T$, $\mathbb{V}$ can be sliced into shorter clips, denoted by $\mathbb{V}[t_1,t_2]$.

To support exploring video clips of different lengths, we organize the video into a multi-level tree structure. The root node of the tree is the entire video, \textit{i.e.}, $\mathbb{V}\equiv\mathbb{V}[0,T]$. The tree has $D$ levels (the root is the $0$-th level and the leaf node is the $D$-th level); each non-leaf node has $K$ children, corresponding to its video clip partitioned into $K$ equal-length, non-overlapping sub-clips. We denote a $d$-th-level clip as $\mathbb{V}_{k_1,\ldots,k_d}$, where $k_{d'}\in\{0,\ldots,K-1\}$ indicates the child index at the $d'$-th level. Unless otherwise specified, we assume that $D=3$ and $K=\mathrm{round}(\sqrt[D]{T/16\mathrm{s}})$ so that the video clip at the leaf level is approximately $16$-second long. This hierarchical structure allows the agent to check long video clips first and, when necessary, `zoom in' to find an answer in finer-scale visual content. While the uniform partition is easy to implement, we understand that it is not the optimal choice, \textit{e.g.}, it would cause semantically similar content to fall into neighboring sub-clips, increasing the ambiguity of localization.

\name is a large reasoning model (LRM) and follows a chain-of-thought-with-tool (CoTwT) framework, where two multimodal tools are incorporated:
\begin{itemize}
\item The \textbf{video captioning} tool, $\mathtt{video\_cap}()$. It receives a video clip $\mathbb{V}_{k_1,\ldots,k_d}$ with the number of sampled frames $F$, and outputs the text description $\mathbf{t}$ of the clip.
\item The \textbf{video QA} tool, $\mathtt{video\_qa}()$. It receives a video clip $\mathbb{V}_{k_1,\ldots,k_d}$ with the number of sampled frames $F$, the question $\mathbf{q}$, and outputs the answer $\mathbf{a}$ (it is possible to answer `I don't know'). This tool is allowed only on the lowest-level clips.
\end{itemize}
There is a major difference between these two tools: $\mathtt{video\_cap}()$ aims to offer generic video descriptions that assist the subsequent steps for key content localization, while $\mathtt{video\_qa}()$, often called at the last step, focuses on answering the specific question. For simplicity, we assume that both tools sample frames time-uniformly from video data, and vanilla visual encoding (\textit{i.e.}, no compression) is performed on the frames.

\subsection{Chain-of-Thought-with-Tool Procedure}
\label{approach:cotwt}

Based on the above preparation, we formulate the inference of \name into a chain-of-thought-with-tool (CoTwT) procedure, which is widely used in multi-round tool-use tasks, such as Ego-R1~\cite{tian2025ego}. A complete inference episode is written as a chain:
\begin{equation}
\label{eqn:episode}
\mathfrak{E}=[\mathbf{S}_1,\mathbf{S}_2,\ldots,\mathbf{S}_L],
\end{equation}
where each $\mathbf{S}_l$ indicates a step:
\begin{equation}
\label{eqn:step}
\mathbf{S}_l=\left\{
\begin{array}{ll}
(\mathbf{r}_l,\mathbf{t}_l), & \mathrm{if}\quad l<L, \\
(\mathbf{r}_l,\mathbf{a}) & \mathrm{if}\quad l=L,
\end{array}
\right.
\end{equation}
where $\mathbf{r}_l$ is the reasoning statement at the $l$-th step, at the end of which contains information indicating which tool is to be called, and $\mathbf{t}_l$ and $\mathbf{a}$ denote the text description and answer, corresponding to the outputs of $\mathtt{video\_cap}()$ and $\mathtt{video\_qa}()$, respectively. Note that the entire episode contains purely natural language (the multimodal tools are called as external functions), making it easier to (1) adapt to the recent advances of LRMs, and (2) explicitly connect thinking with tool-using towards a transparent inference procedure. The procedure is illustrated in Algorithm~\ref{alg:procedure}.

\begin{algorithm}[t]
\caption{Hierarchical Video Reasoning}
\label{alg:procedure}
\begin{algorithmic}[1]
\REQUIRE Video $\mathbb{V}$, question $\mathbf{q}$, reasoning model $\mathtt{rea}()$, multimodal tools $\mathtt{video\_cap}()$ and $\mathtt{video\_qa}()$
\ENSURE Answer $\mathbf{a}$
\vspace{0.3em}
\STATE Tree depth and width: $D=3$, $K=\mathrm{round}(\sqrt[D]{T/16\mathrm{s}})$
\STATE Get top-level caption $\mathbf{t}_0=\mathtt{video\_cap}(\mathbb{V})$
\STATE Initialize chat history: $\mathfrak{E}=[\mathbf{t}_0]$
\STATE Get first reasoning output: $\mathbf{r}_1=\mathtt{rea}(\mathfrak{E},\mathbf{q})$
\STATE Initialize episode length: $L=1$
\WHILE{$\mathbf{r}_L$ does not contain the answer}
    \STATE Parse: $\mathtt{tool}\in\{\mathtt{video\_cap},\mathtt{video\_qa}\}$,$\mathbb{V}_{k_1,\ldots,k_d}$
    \STATE Call the tool: $\mathbf{t}_{L}=\mathtt{tool}(\mathbb{V}_{k_1,\ldots,k_d})$
    \STATE Update chat history: $\mathfrak{E}\leftarrow\mathfrak{E}+[\mathbf{r}_L,\mathbf{t}_L]$
    \STATE Update reasoning output $\mathbf{r}_{L+1}=\mathtt{rea}(\mathfrak{E},\mathbf{q})$
    \STATE Update episode length: $L\leftarrow L+1$
\ENDWHILE
\RETURN Answer extracted from $\mathbf{r}_L$
\end{algorithmic}
\end{algorithm}

\section{Data Curation}
\label{data}

\subsection{Data Preparation}
\label{data:preparation}

We curate a dataset for training \name. We choose the CG-Bench dataset~\cite{chen2024cg} because it contains clue-grounded QA pairs, \textit{i.e.}, we can supervise the model to localize the key sub-clip(s) before answering the question.

CG-Bench contains 1.2K long-form videos, each of which is paired with a diverse set of QA pairs. We chose 800 videos and the corresponding 5.6K QA pairs for generating CoTwT trajectories.

For each video of CG-Bench, $\mathbb{V}$, we first use the Qwen2.5-VL-72B model~\cite{bai2025qwen2} as the function $\mathtt{video\_cap}()$ to extract its text description. The sampled frames $F$ is set to be $256,128,64,32$, and the suggest length of description (in English words) is $400,400,400,200$ for level index $d=0,1,2,3$, respectively. To guide the LRM to locate sub-clips, we modify the prompt (see Appendix~\ref{subsec:appendix-datagen}) to insert absolute timestamps into the caption.

\begin{figure}[t]
\vspace{-0.5cm}
\centering
\includegraphics[width=1.0\linewidth]{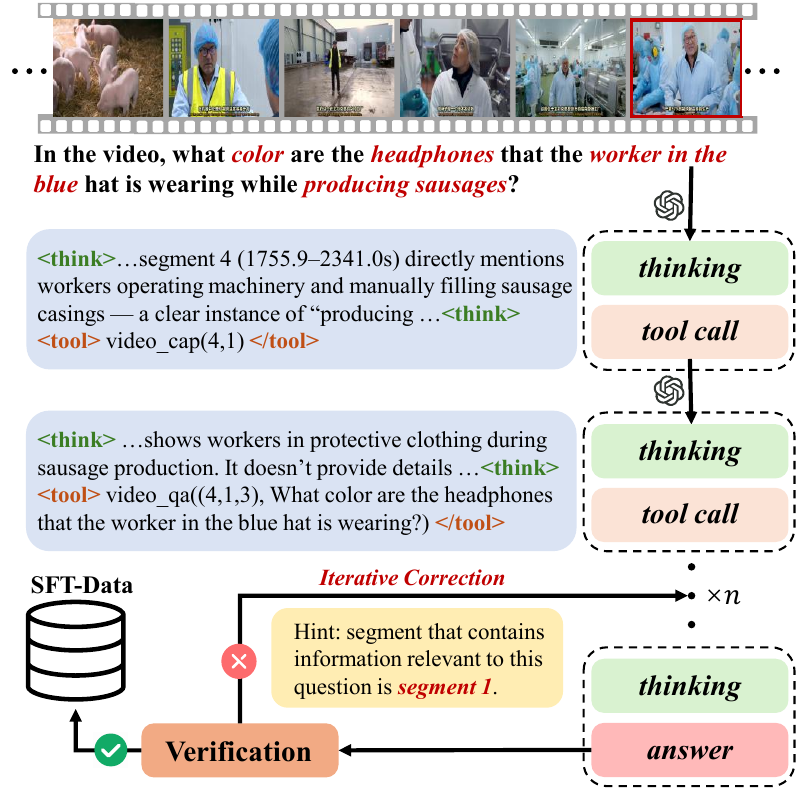}
\caption{An illustration of generating CoTwT trajectories from clue-grounded video QA data. Drawing style was inspired by Ego-R1~\cite{tian2025ego}}.
\label{fig:dataset_pipline}
\end{figure}

\subsection{Generating CoTwT Trajectories}
\label{data:generation}

The pipeline of generating CoTwT trajectories is illustrated in Figure~\ref{fig:dataset_pipline}. We guide GPT-5~\cite{openai2025gpt5} to perform the CoTwT procedure, which starts with the top-level video clip, and continues when the model is confident to produce the final answer. In the prompt to GPT-5 (see Appendix~\ref{subsec:appendix-datagen}), we indicate the functionalities of $\mathtt{video\_cap}()$ and $\mathtt{video\_qa}()$ and specify the rules of using them (\textit{e.g.}, $\mathtt{video\_cap}()$ can be called at a node if its parent has been traversed, and $\mathtt{video\_qa}()$ can only be called in the lowest level nodes).

GPT-5 performs the above task as zero-shot inference; in many (about 30\% of) scenarios, it can produce incorrect answers or fail to pass the above verification. We make two fixes to improve the data quality and guarantee success.
\begin{itemize}
\item Instead of starting with the root node, $\mathbb{V}$, we ask the model to traverse all $K$ sub-clips at the first level. This alleviates the risk that GPT delves deep into local parts without obtaining sufficient global information, and improves stability in particular when exploring hour-long videos.
\item When GPT fails, we use the clue-grounded hints of CG-Bench to guide it towards the correct answer. Meanwhile, we try to keep the hints to a minimal amount: when GPT fails for the first time, we add the highest-level segment containing the relevant event to the prompt; if it still fails, a deeper-level hint with a more precise segment and event description is added. This process continues until the model produces a correct answer. A comparative example of the original and clue-guided prompts is provided in Appendix~\ref{subsec:appendix-datagen}. This strategy guarantees the correctness of each CoTWT trajectory while leaking as few hints as possible. Trained on such data, the LRM learns to generalize toward efficient exploration rather than simply memorizing video content and answers.
\end{itemize}

As a result, we obtain 5.6K CoTwT trajectories with an average of 5.8 steps, yielding approximately 33K high-quality samples for supervised fine-tuning. Upon releasing the SFT data to the community, we show that such data is helpful to train a powerful agent, and the agent's performance is positively related to the amount of SFT data (\textit{i.e.}, it is important to hint GPT when it goes wrong). We have also revealed a promising path to enhance the agent, for which one only needs to establish more clue-grounded video QA pairs and generate more CoTwT trajectories.

\section{Training \name Agent}
\vspace{-0.2cm}
\label{training}

We follow a well-established, two-stage pipeline to train the \name agent, \textit{i.e.}, a supervised fine-tuning (SFT) stage as a cold start and a reinforcement learning (RL) stage for further optimization.

\subsection{Supervised Fine-tuning}
\label{training:sft}

In the first stage, we fine-tune a pretrained large language model on the curated CoTwT data. This cold-start phase equips the model with the ability to generate structured reasoning trajectories under the desired format.

Each training sample simulates a realistic multi-round tool-using process that ultimately leads to the correct answer. Specifically, the reasoning process is enclosed within special tokens $\mathtt{\langle think\rangle}\ldots\mathtt{\mathtt{\langle/ think\rangle}}$, followed by either a tool invocation or an answer. Tool calls are enclosed within $\mathtt{\langle tool\rangle}\ldots\mathtt{\langle /tool\rangle}$ and answers within $\mathtt{\langle answer\rangle}\ldots\mathtt{\langle /answer\rangle}$. During training, the tool invocation content is parsed and executed to obtain corresponding observations, which are then fed back to the model as new contextual information.

This structured annotation enables the model to learn (1) when to continue reasoning, (2) which tool to invoke, and (3) when to terminate reasoning and produce the final answer. After SFT, the model (denoted as \name-SFT) is capable of generating correctly formatted reasoning sequences and performing coherent tool interactions, which serve as a solid foundation for reinforcement learning.

\subsection{Reinforcement Learning with GRPO}
\label{training:rl}

After the SFT stage, we regard the video reasoning process as an interactive exploration environment: the model acts as an agent, video tools ($\mathtt{video\_cap}()$ and $\mathtt{video\_qa}()$) form the action space, and the hierarchical video serves as the environment state. This formulation naturally lends itself to optimization via reinforcement learning.

We employ the GRPO algorithm~\cite{shao2024deepseekmath} with the version
introduced in Ego-R1~\cite{tian2025ego} to further optimize the policy model $\pi_\theta$, aiming to improve reasoning efficiency and accuracy. The objective is defined as:
{\small
\begin{equation}
\begin{aligned}
\mathcal{J}_{\text{GRPO}}(\theta)
&= 
\mathbb{E}_{q \sim P(Q),\, \{o_i^G\}_{i=1}^G \sim \pi_{\theta_{\text{old}}}(O|q)}
\Bigg[
\frac{1}{G} \sum_{i=1}^{G}
\sum_{y=1}^{T}
\frac{1}{|S_i^y|} \sum_{t=1}^{|S_i^y|}
\\[-2pt]
&\quad
\Bigg(
\min \!\Bigg[
\frac{
\pi_\theta (S_{i,t}|q, S_{i,<t})
}{
\pi_{\theta_{\text{old}}}(S_{i,t}|q, S_{i,<t})
}
\hat{A}_{i,t}^y,\;
\\[-2pt]
&\hspace{-4em}
\text{clip}\!\left(
\frac{
\pi_\theta (S_{i,t}|q, S_{i,<t})
}{
\pi_{\theta_{\text{old}}}(S_{i,t}|q, S_{i,<t})
},
1 - \varepsilon, 1 + \varepsilon
\right)
\hat{A}_{i,t}^y
\Bigg)
- \beta\, \mathbb{D}_{\text{KL}}\!\left[\pi_\theta \,\|\, \pi_0\right]
\Bigg)
\Bigg],
\end{aligned}
\end{equation}
}\noindent
\textit{Q} denotes a question sampled from the data distribution \textit{D}, $o_{i}$ represents the model output, \textit{G} is the number of rollouts, and \textit{T} denotes the number of reasoning rounds. The model is parameterized as $\pi_{\theta}$, where $\pi_{\theta}$ and $\pi_{\theta_{\text{old}}}$ denote the current and reference policies, respectively, and $\pi_0$ represents the policy inherited from the \name-SFT model, used for KL regularization. The advantage term is computed by:
\begin{equation}
A_i = 
\frac{
r_i^{\text{GRPO}} - \text{mean}\!\left(\{ r_j^{\text{GRPO}} \}\right)
}{
\text{std}\!\left(\{ r_j^{\text{GRPO}} \}\right)
}.
\end{equation}

\subsection{Reward Design}
\label{training:rewards}

To facilitate the model to explore video content efficiently (while finding the correct answer), we design a composite reward function:
\begin{equation}
\label{eqn:reward}
R = w_\mathrm{ans} \cdot r_{\mathrm{ans}} + w_\mathrm{loc} \cdot r_{\mathrm{loc}} + w_\mathrm{repeat} \cdot r_{\mathrm{repeat}},
\end{equation}
where $w_\cdot$ are reward weights and the three components are defined as follows:
\begin{itemize}
\item The \textbf{answer reward}, $r_{\text{ans}} \in \{0, 1\}$, gives a reward of $1$ if the final answer matches the ground-truth, otherwise $0$.
\item The \textbf{location reward}, $r_{\mathrm{loc}}$, encourages the model to identify the correct segment efficiently:
\[
r_{\mathrm{loc}} = 2 \cdot \frac{\mathrm{cov} \times \mathrm{pre}}{\mathrm{cov} + \mathrm{pre}},
\]
where the coverage and precision are defined as
\[
\mathrm{cov} = \frac{|\mathcal{I}_{\mathrm{model}} \cap \mathcal{I}_{\mathrm{gt}}|}{|\mathcal{I}_{\mathrm{gt}}|}, \quad
\mathrm{pre} = \frac{|\mathcal{I}_{\mathrm{model}} \cap \mathcal{I}_{\mathrm{gt}}|}{|\mathcal{I}_{\mathrm{model}}|},
\]
where $\mathcal{I}_\mathrm{gt}$ and $\mathcal{I}_\mathrm{model}$ indicates the ground-truth and predicted sets of time intervals. $\mathcal{I}_\mathrm{model}$ is the union of all non-overlapping time segments corresponding to the nodes requested by the model. This $F_1$-like metric encourages high coverage of relevant content while penalizing unnecessary exploration.
\item The \textbf{repeat penalty}, $r_{\mathrm{repeat}}$, discourages repeatedly visiting the same segments, reducing wasted computation.
\end{itemize}

\subsection{Rollout and Optimization}
\label{training:rollout}

During RL training, the agent interacts with executable video tools to generate rollout trajectories. Each rollout continues until the model outputs a final answer or reaches a predefined maximum number of reasoning steps. The collected trajectories are then used to compute policy gradients and update $\pi_\theta$ using GRPO.

After RL training, the resulting model, \name, is capable of performing multi-tool reasoning efficiently on long video tasks. It learns to minimize redundant exploration while maintaining high answer accuracy, achieving a superior tradeoff between performance and computational efficiency compared to conventional MLLMs.

\begin{table*}[!t]
\vspace{-0.5cm}
\small
\centering
\renewcommand{\arraystretch}{0.98}
\setlength{\tabcolsep}{2pt}
\begin{minipage}{0.48\linewidth}
\centering
\resizebox{\textwidth}{!}{
\begin{tabular}{lccccccc}
\toprule
\textbf{Method} & ER & EU & KIR & TG & Rea & Sum & \textbf{Overall} \\
\midrule
\multicolumn{8}{l}{\textit{Proprietary Models}} \\
Gemini 1.5 Pro~\cite{team2024gemini} & 32.1 & 30.9 & 39.3 & 31.8 & 27.0 & 32.8 & 33.1 \\
GLM-4V-plus-0111~\cite{hong2025glm} & 46.2 & 47.8 & 54.1 & 42.7 & 46.5 & 37.9 & 48.7 \\
GPT-4o-20241120~\cite{hurst2024gpt} & 48.9 & 49.5 & 48.1 & 40.9 & 50.3 & 50.0 & 48.9 \\
\midrule
\multicolumn{8}{l}{\textit{Leading Open-sourced MLLMs}} \\
TimeMarker-8B~\cite{chen2024timemarker} & 42.8 & 39.1 & 34.9 & 38.7 & 38.2 & 48.8 & 41.3 \\
VideoLLaMA3-7B~\cite{zhang2025videollama} & 45.8 & 42.4 & 47.8 & 35.9 & 45.8 & 36.2 & 45.3 \\
InternVL2.5-78B~\cite{chen2024expanding} & 43.8 & 42.0 & 42.1 & 36.8 & 51.0 & 37.9 & 43.6 \\
Qwen2-VL-72B~\cite{bai2025qwen2} & 38.0 & 41.1 & 38.3 & 41.4 & 46.5 & 46.6 & 41.3 \\
ReTake-7B~\cite{wang2024retake} & 49.8 & 46.2 & 52.9 & 45.0 & 45.8 & 27.6 & 47.8 \\
VideoChat-Flash-7B~\cite{li2025videochat} & 51.1 & 46.0 & 49.0 & 38.9 & 48.5 & 34.5 & 48.2 \\
AdaReTake-72B~\cite{wang2025adaretake} & 53.0 & 50.7 & 62.2 & 45.5 & 54.7 & 37.9 & 53.3 \\
\midrule
\multicolumn{8}{l}{\textit{Agent-based Systems}} \\
VideoAgent~\cite{wang2024videoagent} & 28.0 & 30.3 & 28.0 & 29.3 & 28.0 & 36.4 & 29.3 \\
VideoTree~\cite{wang2025videotree} & 30.3 & 25.1 & 26.5 & 27.7 & 31.9 & 25.5 & 28.8 \\
MemVid~\cite{yuan2025memory} & 53.4 & 40.6 & 37.8 & 43.9 & 43.2 & 28.1 & 44.4 \\
VCA~\cite{yang2025vca} & 43.7 & 40.8 & 37.8 & 38.0 & 46.2 & 27.3 & 41.3 \\
\midrule
\rowcolor{gray!15}
\name & 49.2 & 48.4 & 56.4 & 56.4 & 44.3 & 43.1 & 50.0 \\
\rowcolor{red!15}
\name (new\textsuperscript{$\dagger$}) & 60.9 & 57.8 & 70.1 & 62.7 & 50.2 & 55.2 & 60.7 \\
\bottomrule
\end{tabular}
}
\caption{QA accuracy (\%) on all sub-tasks of LVBench~\cite{wang2025lvbench}. \textsuperscript{$\dagger$} We trained an updated version using video captions generated by Qwen3-VL-32B-Instruct and renewed SFT data.}
\label{tab:lvbench}
\end{minipage}
\hfill
\begin{minipage}{0.22\linewidth}
\centering
\resizebox{\linewidth}{!}{
\renewcommand{\arraystretch}{1.05}
\begin{tabular}{lc}
\toprule
\textbf{Method} & \textbf{Acc.} \\
\midrule
\multicolumn{2}{l}{\textit{Proprietary Models}} \\
GPT-4V~\cite{hurst2024gpt} & 49.2 \\
GPT-4o~\cite{hurst2024gpt} & 64.6 \\
\midrule
\multicolumn{2}{l}{\textit{Open-Sourced MLLMs}} \\
Video-CCAM-14B~\cite{fei2024video} & 63.1 \\
Video-XL-7B~\cite{shu2025video} & 64.9 \\
LLaVA-OV-72B~\cite{li2024llava} & 66.4 \\
LinVT-7B~\cite{gao2024linvt} & 68.9 \\
Aria-25B~\cite{li2024aria} & 70.6 \\
Oryx-1.5-32B~\cite{liu2024oryx} & 72.3 \\
VideoLLaMA3-7B~\cite{zhang2025videollama} & 73.0 \\
VideoChat-Flash-7B~\cite{li2024videochat} & 74.7 \\
\midrule
\multicolumn{2}{l}{\textit{Agent-based Systems}} \\
VideoTree~\cite{wang2025videotree} & 60.4 \\
VideoMind-7B~\cite{liu2025videomind} & 64.4 \\
\midrule
\rowcolor{gray!15}
\name & 68.1 \\
\rowcolor{red!15}
\name (new\textsuperscript{$\dagger$}) & 71.3 \\
\bottomrule
\end{tabular}
}
\caption{Model-level QA accuracy (\%) on the MLVU dataset~\cite{zhou2025mlvu}. \textsuperscript{$\dagger$} We trained an updated version using the same setting as in Table~\ref{tab:lvbench}.}
\label{tab:mlvu}
\end{minipage}
\hfill
\begin{minipage}{0.28\linewidth}
\centering
\resizebox{\linewidth}{!}{
\renewcommand{\arraystretch}{0.85}
\setlength{\tabcolsep}{4pt}
\begin{tabular}{l c c}
\toprule
\textbf{Method} & \textbf{w/o} & \textbf{w/} \\
\midrule
\multicolumn{3}{l}{\textit{Proprietary Models}} \\
Gemini 1.5 Flash~\cite{team2024gemini} & 61.1 & 68.8 \\
GPT-4o-20240513~\cite{hurst2024gpt} & 65.3  & 72.1 \\
Gemini 1.5 Pro~\cite{team2024gemini} & 67.4 & 77.4 \\
\midrule
\multicolumn{3}{l}{\textit{Leading Open-sourced MLLMs}} \\
Qwen2-VL-72B~\cite{wang2024qwen2} & 62.2 & 74.3\\
InternVL2.5-72B~\cite{chen2024expanding} & 62.6 & 64.8 \\
LLaVA-OV-72B~\cite{li2024llava} & 60.0 & 62.4 \\
LLaVA-Video-72B~\cite{zhang2024video} & 61.5 & 72.5\\
VideoChat-Flash-7B~\cite{li2024videochat} & 55.6 & 63.3\\
VideoLLaMA3-7B~\cite{zhang2025videollama} & 54.9 & 61.0\\
LiveCC-7B~\cite{chen2025livecc} & 53.7 & 64.1 \\
\midrule
\multicolumn{3}{l}{\textit{Agent-based Systems}} \\
VideoAgent~\cite{wang2024videoagent} & 46.4 & -- \\
VideoTree~\cite{wang2025videotree} & 54.2 & -- \\
MemVid~\cite{yuan2025memory} & 55.0 & -- \\
Video-RAG-7B~\cite{ren2025videorag} & -- & 59.8 \\
Ego-R1~\cite{tian2025ego} & -- & 64.9 \\
VCA~\cite{yang2025vca} & 56.3 & -- \\
\midrule
\rowcolor{gray!15}
\name & 55.8 & 64.4 \\
\rowcolor{red!15}
\name (new\textsuperscript{$\dagger$}) & 58.0 & 68.6 \\
\bottomrule
\end{tabular}
}
\caption{QA accuracy (\%) on the `long' subset of Video-MME~\cite{fu2025video} without (w/o) or with (w/) subtitles. \textsuperscript{$\dagger$} We trained an updated version using the same setting as in Table~\ref{tab:lvbench}.}
\label{tab:videomme}
\end{minipage}
\end{table*}

\section{Experiments}
\label{experiments}

\subsection{Implementation Details}
\label{experiments:details}

We train \name upon a Qwen3-8B model~\cite{yang2025qwen3}. The multimodal tools, $\mathtt{video\_cap}()$ and $\mathtt{video\_qa}()$, are chosen to be Qwen2.5VL-72B and Qwen2.5VL-32B~\cite{bai2025qwen2}, respectively. Compared to other agentic approaches~\cite{tian2025ego,yang2025vca,wang2025videotree,wang2024videoagent}, that relied on proprietary LLMs such as GPT or Gemini, our setting eases local deployment and fair comparison. We perform SFT for 3 epochs followed by RL for 2 epochs.

\name is tested on three popular long-form video understanding benchmarks. LVBench~\cite{wang2025lvbench} contains 103 videos (average duration: 4038 seconds) and 1,549 QA pairs. Video-MME-long~\cite{fu2025video} contains 300 videos (average duration: 41 minutes) with 3 QA pairs for each video. MLVU~\cite{zhou2025mlvu} contains 1,337 videos with their durations ranging between 3 minutes to 2 hours. All these benchmarks have provided multiple choices for each question; we prompt these choices with questions to \name and ask it to produce the choice index in the answer box. An answer is considered correct if the choice(s) perfectly match the ground-truth.

\subsection{Results and Analysis}
\label{experiments:results}

\noindent\textbf{Results on LVBench.}
We compare \name with state-of-the-art models on LVBench in Table~\ref{tab:lvbench}. The compared methods are categorized into three groups: proprietary models, leading open-sourced MLLMs, and agent-based systems. As illustrated in Table~\ref{tab:lvbench}, \name achieves a $50.0\%$ accuracy, outperforming the other agent-based methods by at least $5.6\%$. Besides, with an 8B-LLM, \name surpasses most proprietary and open-sourced MLLMs; for example, it exceeds GPT-4o by 1.1\% and GLM-4V-plus by 1.3\%. Notably, \name demonstrates outstanding results on two sub-categories, KIR (Key Information Retrieval) and TG (Temporal Grounding) tasks. In particular, its performance on TG reaches $56.4\%$, surpassing all other models by a significant margin of $10.9\%$. These results highlight the strong ability of \name in accurately locating key temporal segments within long videos. Moreover, the ability of \name grows with the multimodal tools: as shown in Table~\ref{tab:lvbench}, when we use Qwen3-VL-32B-Instruct, a stronger MLLM, for video captioning, the overall accuracy improves significantly, meanwhile the advantages in the KIR and TG sub-categories persist.

\noindent\textbf{Results on MLVU and Video-MME.}
The comparisons are shown in Tables~\ref{tab:mlvu} and~\ref{tab:videomme}, respectively. While \name also performs well, it does not excel among the open-sourced MLLMs. The reason lies in the property of the benches: MLVU contains many short videos, and Video-MME contains many global questions like `\textit{What is the main idea of the video?}', which is beneficial for the uniform or adaptive (\textit{e.g.}, \cite{tang2025adaptive}) frame sampling methods. \name's advantage also reflects in inference time. We compare \name with Ego-R1~\cite{tian2025ego}, which reports a similar accuracy on Video-MME. Differently, Ego-R1 requires video captioning every $30$ seconds, resulting in an average of $86$ caption segments on VideoMME, while \name only undergoes an average of $10.5$ rounds, claiming a much lower computational cost. The model's performance on MLVU and Video-MME also benefits from improved multimodal tools, \textit{e.g.}, Qwen3-VL-32B-Instruct for video captioning.

\begin{table}[t]
\vspace{-0.3cm}
\centering
\renewcommand{\arraystretch}{0.85}
\setlength{\tabcolsep}{4pt}
\begin{tabular}{lcccc}
\toprule
\textbf{Method} & KIR & TG & \textbf{LVBench} & \textbf{V-MME/L} \\
\midrule
SFT$_\mathrm{10K}$      & 46.4 & 39.1 & 39.1 & 57.7\\
SFT$_\mathrm{full}$         & 48.5 & 50.9 & 41.6 & 59.2\\
Ours$_\mathrm{10K}$ w/ RL    & 54.0 & 55.0 & 47.4 & 60.2\\
Ours$_\mathrm{full}$ w/ RL  & 56.4 & 56.4 & 50.0 & 64.4 \\
\bottomrule
\end{tabular}
\caption{QA accuracy (\%) with respect to SFT data size.}
\label{tab:ablation_sft_scale}
\end{table}

\begin{table}[t]
\vspace{-0.3cm}
\centering
\renewcommand{\arraystretch}{0.85}
\setlength{\tabcolsep}{4pt}
\begin{tabular}{lcccc}
\toprule
\textbf{Method} & KIR & TG & \textbf{LVBench} & \textbf{V-MME/L} \\
\midrule
SFT         & 48.5 & 50.9 & 41.6 & 59.2 \\
w/o $r_\mathrm{loc}$ & 49.1 & 53.2 & 45.8 & 61.4\\
Ours    & 56.4 & 56.4 & 50.0 & 64.4\\
\bottomrule
\end{tabular}
\caption{QA accuracy (\%) with or without the location reward.}
\label{tab:ablation_method}
\end{table}

\begin{table}[t]
\vspace{-0.2cm}
\centering
\renewcommand{\arraystretch}{0.85}
\setlength{\tabcolsep}{1pt}
\begin{tabular}{lccccc}
\toprule
\textbf{Model} & KIR & TG & \textbf{LVBench} & \textbf{V-MME/L} & \textit{Time} \\
\midrule
Qwen2.5-VL-3B          & 48.8 & 53.2 & 44.5 & 56.0 & 50.5\\
Qwen2.5-VL-7B & 51.5 & 52.7 & 46.0 & 57.8 & 57.4\\
Qwen2.5-VL-32B   & 58.4 & 54.5 & 49.8 & 63.9 & 135.0\\
Qwen2.5-VL-72B   & 56.4 & 56.4 & 50.0 & 64.4 & 175.7\\
\bottomrule
\end{tabular}
\caption{QA accuracy (\%) with respect to MLLM scales. The last column shows average inference time (on LVBench, in seconds).}
\label{tab:ablation_diff_model}
\end{table}

\begin{table}[t]
\vspace{-0.2cm}
\centering
\renewcommand{\arraystretch}{0.85}
\setlength{\tabcolsep}{2pt}
\begin{tabular}{lccccc}
\toprule
\textbf{Max Rounds} & KIR & TG & \textbf{LVBench} & \textbf{V-MME/L} & \textit{Time (s)} \\
\midrule
10   & 49.1 & 52.3 & 43.0 & 57.1 & 103.6\\
20   & 49.5 & 58.2 & 47.3 & 63.2 & 170.8\\
25   & 54.6 & 54.1 & 48.5 & 63.6 & 172.8\\
30   & 56.4 & 56.4 & 50.0 & 64.4 & 175.7\\
\bottomrule
\end{tabular}
\caption{QA accuracy (\%) with respect to maximum rounds of tool use (mainly the $\mathtt{video\_cap}()$ function).}
\label{tab:ablation_diff_rounds}
\end{table}

\begin{figure*}
\centering
\includegraphics[width=0.99\linewidth]{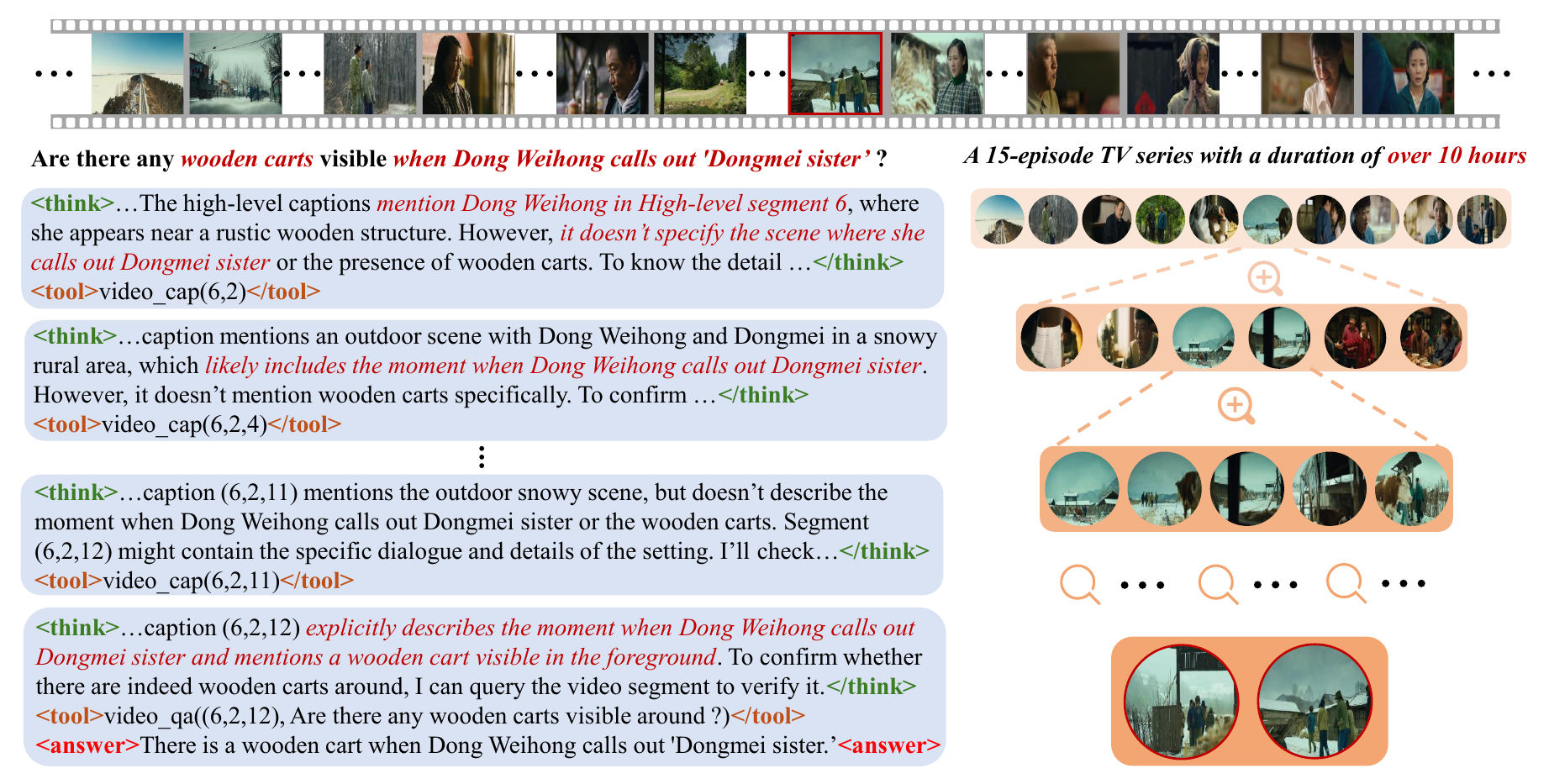}
\caption{\name can navigate in ultra-long videos efficiently. We show an example in a long-form TV drama, \textit{A Lifelong Journey}.}
\label{fig:ultra_long_videos}
\end{figure*}

\noindent\textbf{Ablative studies.}
We ablate the performance with respect to the SFT and RL strategies on the LVBench dataset. As shown in Table~\ref{tab:ablation_sft_scale}, the model fine-tuned with the full 33K SFT samples outperforms the one trained with a subset of 10K samples, both after the SFT and subsequent RL stages. This demonstrates the importance of increasing the size of SFT data. Another important factor is the location reward, $r_\mathrm{loc}$. As shown in Table \ref{tab:ablation_method}, adding $r_{\mathrm{loc}}$ leads to significant performance gains in the overall set and the KIR (Key Information Retrieval) and TG (Temporal Grounding) subsets. These results indicate that $r_{\mathrm{loc}}$ effectively enhances the model's ability of video navigation and such ability contributes to long video QA.

\noindent\textbf{Accuracy-efficiency tradeoff.}
Figure~\ref{fig:teaser} shows that \name achieves a favorable tradeoff; it achieves a $50.0\%$ accuracy on LVBench, requiring $3$ minutes per QA. The cost can be reduced to $2$ minutes per QA at a mere $0.2\%$ accuracy drop. More results are shown in Tables~\ref{tab:ablation_diff_model} and~\ref{tab:ablation_diff_rounds}, where we change the MLLM scale for $\mathtt{video\_cap}()$ and alter the maximum number of tool uses, respectively. These results suggest an interesting solution to further improve the tradeoff, \textit{i.e.}, switching the setting of tool use (in both size and number) to answer questions of different difficulties.

\begin{figure*}[!t]
\centering
\includegraphics[width=0.99\linewidth]{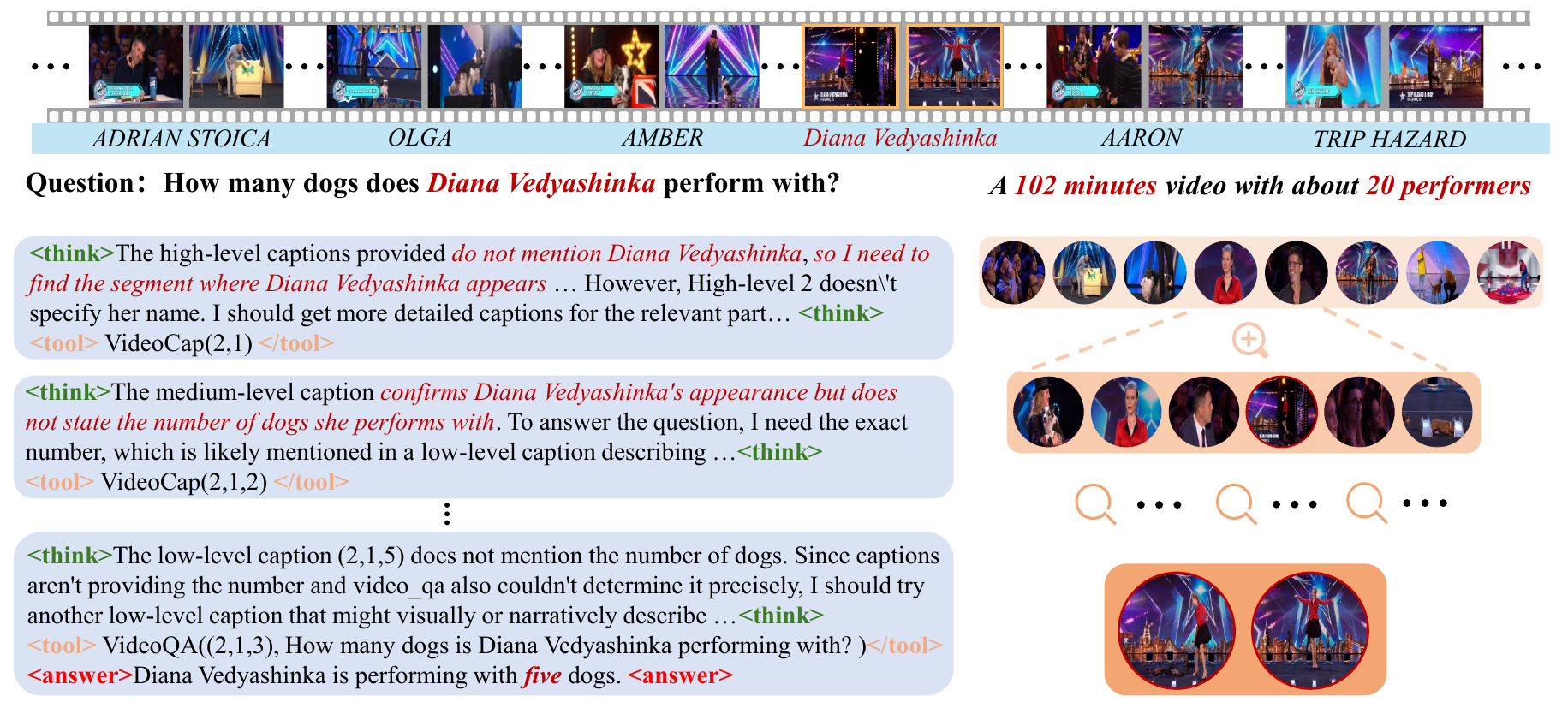}
\caption{An example of how \name smartly navigates to the critical segment and answers the question.}
\label{fig:qualitative_example}
\end{figure*}

\noindent\textbf{Case studies.}
We further conduct a case study to illustrate the reasoning and planning capability of \name. The video in Figure~\ref{fig:qualitative_example} is $102$ minutes long and features around $20$ performers. The question requires the model to identify a specific performer and then accurately count the number of dogs associated with that performer. Without prior information about the performer, \name first locates the segment where the target performer appears, after which it explores fine-grained sub-segments to pinpoint the exact moment of the performance. Finally, it invokes the video QA tool to obtain the precise answer. This example demonstrates the strong reasoning, planning, and temporal localization abilities of \name for efficient long-video understanding.

However, \name can sometimes be distracted by other segments that are semantically related to the question. As shown in the Appendix \ref{sec:appendix-failure}, the model may get stuck in an irrelevant segment instead of shifting its focus to the correct one. In contrast, humans can easily recognize such errors and redirect attention to the appropriate segment. Interestingly, we find that providing simple textual hints can effectively guide LongVideo-R1 back to the correct segment, enabling it to produce the right answer.

\subsection{Extension to Ultra-long Videos}

Beyond existing benchmarks, \name also excels in ultra-long video QA. As illustrated in Figure~\ref{fig:ultra_long_videos}, \name smartly navigates to the accurate location (and gets the correct answer) with $10$--$20$ rounds, even when the input video is tens of hours long. In comparison, open-sourced MLLMs (even sampling $256$ frames) can barely find efficient information for QA, and other agent-based systems like Ego-R1~\cite{tian2025ego} and VideoTree~\cite{wang2025videotree} require the number of samples to grow linearly with video duration, leading to prohibitively high computational costs.

\subsection{Future Directions}
\vspace{-0.1cm}
\label{experiments:directions}

Our work, a preliminary study towards low-cost long video understanding, reveals a few research directions for the future.
\begin{itemize}
\item \textbf{Extended tools.} \name only considered two tools (besides reasoning), $\mathtt{video\_cap}()$ and $\mathtt{video\_qa}()$. In the future, one may introduce more tools (\textit{e.g.}, video instance recognition, video clip segmentation) to further improve the model's ability. In such scenarios, an extra reward term shall be added to penalize the aggregated computational cost of tool use.
\item \textbf{Advanced settings.} We assumed that each video QA is processed individually. In practice, if one video corresponds to multiple QA pairs, the best model choice may vary, \textit{e.g.}, the model can spend more time on key information indexing because the overhead can be amortized among all pairs. There can also emerge related settings, like incremental QA, that require the model to reuse the information efficiently.
\item \textbf{Enhanced video descriptions.} \name was built upon an LRM whose performance heavily relies on quality video captions. It hence emerges a new topic -- enhancing the video description tools for more accurate and efficient reasoning and navigation. We look forward to the agent and tools being optimized simultaneously in a unified framework.
\end{itemize}

\section{Conclusion}
\vspace{-0.1cm}
\label{conclusions}

This paper presents \name, an agentic framework designed for efficient long video understanding. \name explores long videos like humans: starting with top-level video sections, it maintains video descriptions and performs reasoning to judge whether the question can be answered and which part of video is to be navigated next. \name was fine-tuned upon a pre-trained LRM via SFT (on a curated dataset) and RL, and we show that richer SFT data can help. \name achieves competitive QA accuracy on several long video benchmarks and is skilled at information retrieval and grounding tasks; more importantly, it shows a favorable accuracy-efficiency tradeoff over other agentic algorithms. We hope \name enlightens new directions for long video understanding.

{
    \small
    \bibliographystyle{ieeenat_fullname}
    \bibliography{main}
}

\clearpage
\appendix
The supplementary document provides (1) details of our hierarchical video definition in \ref{sec:appendix-def};(2)comprehensive implementation details including the prompts we used for data generation and the experiment setup in \ref{sec:appendix-detail};(3)more qualitative examples and ultra long video examples in \ref{sec:appendix-examples};(4)failure examples and analysis in \ref{sec:appendix-failure}.

\section{Hierarchical Video Definition}
\label{sec:appendix-def}
To enable efficient localization of relevant segments within long videos without requiring any preprocessing, we represent each video using a hierarchical tree structure. Given a tree depth $D$ and width $W$, the root node corresponds to the entire video. The root is evenly divided into $W$ child segments, each of which is recursively divided into another $W$ sub-segments. Repeating this process yields a hierarchical video tree with $D$ levels.

In practice, we set $D=3$, and choose $W$ adaptively according to video length. The leaf-level segment length is fixed at $16$ seconds. Let \textit{Duration} denote the total video length, then the number of leaf segments is $\textit{Duration}/16$. We determine the width as:
\[
W = \left(\frac{\mathrm{Duration}}{16}\right)^{1/D},
\]
which typically lies within $4$ to $8$ across all datasets.

To ensure that deeper nodes receive fine-grained visual signals, we adjust both frame sampling rate and spatial resolution according to the hierarchy. For the caption model, we use \{256, 128, 64, 32\} frames from level 0 to level 3, respectively. The corresponding image resolutions are set to 
\[
\frac{512}{2\sqrt{2}},\;\frac{512}{2},\;\frac{512}{\sqrt{2}},\;512,
\]
ensuring that the overall number of visual tokens remains approximately constant across levels. This design guarantees consistent computation cost per caption call while enabling finer temporal and spatial detail as the model traverses downward in the hierarchy.

\section{Implementation Details}
\label{sec:appendix-detail}
\subsection{Environment Setup}
\label{subsec:appendix-setup}
During both data generation and model evaluation, we use Qwen-2.5-VL-72B as our \textit{video\_cap} model and Qwen-2.5-VL-32B as our \textit{video\_qa} model. These two components can be replaced by other models; for instance, Qwen-2.5-VL-32B is also capable of serving as an effective captioning model.

For CoTWT data generation, we employ GPT-5 as the central reasoning model. All SFT and RL training is conducted on a cluster with 8 NVIDIA H800 GPUs (80GB). Mixed-precision training and FSDP sharding are used to maximize training throughput.

\begin{table}[h]
\centering
\setlength{\tabcolsep}{0.08cm}
\begin{tabular}{lcc}
\toprule
\textbf{Configuration} & SFT & RL  \\
\midrule
Model init     & Qwen3-8B & LongVideo-R1-SFT \\
LLM sequence length        & 32768 & 32768 \\
Learning rate    & $1\times 10^{-5}$ & $1\times 10^{-6}$ \\
Learning rate schedule  & cosine decay & constant\\
Training epochs  & 3 & 2\\
Global batch size & 32 & 12\\
Training steps & 384 & 696\\
Rollout numbers & - & 16\\
\bottomrule
\end{tabular}
\caption{Training hyper-parameters.}
\label{tab:hyper-para}
\end{table}

\subsection{Data Generation}
\label{subsec:appendix-datagen}

We use a proprietary GPT-5 model to generate the CoTWT supervision signals. Since CGBench provides timestamp annotations for each question, we use CGBench as the primary source for CoTWT construction. CGBench contains approximately 1200 videos and 12,000 QA pairs. We use 800 videos and around 8000 QA pairs to construct the SFT data; after filtering, we obtain 5600 high-quality CoTWT trajectories.

The remaining 400 videos and approximately 4200 QA pairs are reserved for RL training.

The prompts used for data generation and caption extraction are listed in Table \ref{tab:system_prompt} and Table \ref{tab:video_cap_prompt} of the appendix. Initially, we provided only the root-level caption as GPT-5’s starting context, but this resulted in unstable behavior and low accuracy (around 30\%). We found that providing the $W$ child captions from the highest-level nodes as initial information substantially improves stability and accuracy.

For all datasets (CGBench, LVBench, VideoMME-Long), the tree width $W$ is set between 4 and 8. For EgoSchema, due to its large number of short 2-minute videos, we set $W$ between 3 and 8.

\begin{table*}[h!]
\centering
\caption{System prompt for data generation.}
\label{tab:system_prompt}
\begin{minipage}{0.99\linewidth}
\begin{tcolorbox}
\textbf{[BEGIN OF GOAL]}\\
You are a reasoning assistant designed to answer questions about a long video through hierarchical captions. 
The video is organized into three levels of temporal granularity: \\
1. High-level: The video is divided into \textbf{width} major segments. \\
2. Medium-level: Each High-level segment is further divided into \textbf{width} sub-segments. \\
3. Low-level: Each Medium-level segment is further divided into \textbf{width} finer sub-segments.\\
You will be asked a question about the video.\\
At the beginning, you are given **only the High-level captions**.\\
Your goal is to answer the question as accurately as possible.\\
\textbf{[END OF GOAL]}

\vspace{6pt}

\textbf{[BEGIN OF REASONING AND TOOL USAGE INSTRUCTIONS]}\\
1. Reason first:\\ 
   Before taking any action, carefully analyze whether the current information (captions you already have) is sufficient to answer the question.\\
2. If sufficient:\\ 
   Directly provide your final answer inside $\mathtt{\langle answer\rangle}\mathtt{\langle /answer\rangle}$ tags.\\
3. If insufficient:\\  
   Identify which part(s) of the video might contain the needed information.  
   Then use one of the following tools:\\
   - To obtain finer captions:\\
     $\mathtt{\langle tool\rangle}$get\_caption((high\_segment\_id, medium\_segment\_id, low\_segment\_id))$\mathtt{\langle /tool\rangle}$\\
     - Each of the three IDs is an integer from 1 to \textbf{width}.\\
     - To request a Medium-level caption, provide (high\_segment\_id, medium\_segment\_id) only.\\
     - To request a Low-level caption, provide the full triplet (high\_segment\_id, medium\_segment\_id, low\_segment\_id).\\
   - To query visual information from the actual video segment:\\
     $\mathtt{\langle tool\rangle}$video\_qa((high\_segment\_id, medium\_segment\_id, low\_segment\_id), query)$\mathtt{\langle /tool\rangle}$\\
     - This tool sends the corresponding Low-level video segment to a specialized video QA module.  \\
     - The query should specify what exact information you need (e.g., “what color is the person’s shirt?”, “what object is on the table?”). \\ 
     - You may only use video\_qa after you have already retrieved the corresponding Low-level caption for that segment.\\
4. Restriction:\\
   In each reasoning round, you may only call one tool (either `get\_caption` or `video\_qa`) once to obtain new information.\\
\textbf{[END OF REASONING AND TOOL USAGE INSTRUCTIONS]}

\vspace{6pt}

\textbf{[BEGIN OF FORMAT INSTRUCTIONS]}\\
Your reasoning and actions must follow this structure exactly:
$\mathtt{\langle think\rangle}$Your internal reasoning process here. Analyze what information you have, what is missing, and which part might be relevant.$\mathtt{\langle /think\rangle}$
$\mathtt{\langle tool\rangle}$(get\_caption or video\_qa call here, if needed)$\mathtt{\langle /tool\rangle}$ 
or  
$\mathtt{\langle think\rangle}\ldots\mathtt{\mathtt{\langle/ think\rangle}}$
$\mathtt{\langle answer\rangle}$Your final answer here (only when you are confident the information is sufficient).$\mathtt{\langle /answer\rangle}$\\
\textbf{[END OF FORMAT INSTRUCTIONS]}

\end{tcolorbox}
\end{minipage}
\end{table*}

\begin{table*}[t]
\centering
\caption{Video Caption Model Prompt.}
\label{tab:video_cap_prompt}
\begin{minipage}{0.99\linewidth}
\begin{tcolorbox}
    You are a video understanding expert. Please create a detailed description with timestamp information for the current video clip (which contains multiple frames arranged in chronological order).\\
    You are given \textbf{num\_frame} uniformly sampled frames from the video, along with the timestamp (in seconds) of each frame in the entire video.\\
    Description Guidelines:\\
    -Dialogue Description Guidelines:\\
    1)In addition to video frames, subtitle information for this video segment is also provided.\\
    2)The output description must faithfully include the given subtitle content. Do not invent or add dialogues that are not provided. Avoid redundant repetition, maintain the original order of the lines, and ensure the sentences flow smoothly.\\
    3)Your output should be around \textbf{num\_words} words.\\
    -Whenever reasonable, include the provided timestamps in your description.\\
    1)For multiple frames with short intervals that depict the same continuous action, you may merge them into a single description.\\
    2)For example:This video begins at 0.0s with a scene featuring two individuals seated outdoors, engaging in a conversation. The subtitles indicate they are discussing the impact of the pandemic on their ability to shoot videos at a bar. By 14.0s, the dialogue shifts to their newfound regular appearances on a show called Scam Nation. At 28.0s, the conversation turns to the promotion of a product named Kraken, encouraging viewers to visit a website for...\\
    Output Format:\\
    Your response should be in the following format, wrapped with $\mathtt{\langle caption\rangle}\mathtt{\langle /caption\rangle}$ tags: "$\mathtt{\langle caption\rangle}$This clip (video) XXX$\mathtt{\langle /caption\rangle}$".
\end{tcolorbox}
\end{minipage}
\end{table*}

\subsection{Training hyper-parameters}
\label{subsec:appendix-trainpara}

We adopt Qwen3-8B as the central reasoning model for both SFT and RL training. The model receives the $W$ highest-level captions as its initial observation and interacts with the video tree through a sequence of tool calls.

Training consists of two phases:

\paragraph{SFT.}  
We train the model to imitate CoTWT trajectories by predicting reasoning process, search actions and answers. This stage helps the model acquire hierarchical search behavior and structured video reasoning skills.

\paragraph{RL.}  
During RL, we pre-extract all hierarchical captions to accelerate training, while the \textit{video\_qa} tool is invoked in real time. Qwen-2.5-VL-32B is deployed on two GPUs to serve as the \textit{video\_qa} module, while the remaining six GPUs are dedicated to RL training.

The detailed hyper-parameters are listed in Table \ref{tab:hyper-para}.

\subsection{Time consumption calculation}
The timing was tested on an A800. The total inference time of LongVideo-R1 consists of three components: (1) the forward-pass time of the reasoning model $T_1$, (2) the captioning time required for processing a video segment $T_2$, and (3) the time required for the video\_qa model to answer a query $T_3$. Let the average number of calls to LongVideo-R1, the video\_cap model, and the video\_qa model be $C_1$, $C_2$, and $C_3$, respectively. Then the expected time cost for answering one question is:
\[
T = C_1 T_1 + C_2 T_2 + C_3 T_3.
\]

Using VideoMME-Long as an example, the model requires on average 10.5 reasoning rounds per question. Therefore, $C_1 = 10.5$. The video\_qa model is invoked infrequently, with an average of $C_3 = 0.36$ calls per question.

Since the average tree width for VideoMME-Long is $W = 5$, the initial step requires obtaining the $W$ highest-level captions. During the subsequent reasoning process, every reasoning step except the one that triggers a video\_qa call requires an additional caption. Thus the expected number of caption calls is:
\[
C_2 = W + C_1 - 1 - C_3 = 5 + 10.5 - 1 - 0.36 = 14.14.
\]

Assuming Qwen-2.5-VL-32B is used for both video\_cap and video\_qa, the empirical average runtimes are:
\[
T_1 \approx 2.5\text{s}, \quad T_2 \approx 7.0\text{s}, \quad T_3 \approx 2.7\text{s}.
\]

Therefore, the expected end-to-end time required to answer a single question on VideoMME-Long is:
\[
T = 10.5 \times 2.5 + 14.14 \times 7.0 + 0.36 \times 2.7 
    \approx 135\text{s}.
\]

This result reflects the full hierarchical search procedure, including both caption retrieval and occasional fine-grained Video\_QA queries.

\section{More Qualitative Examples}
\label{sec:appendix-examples}

\begin{figure*}
\centering
\includegraphics[width=0.99\linewidth]{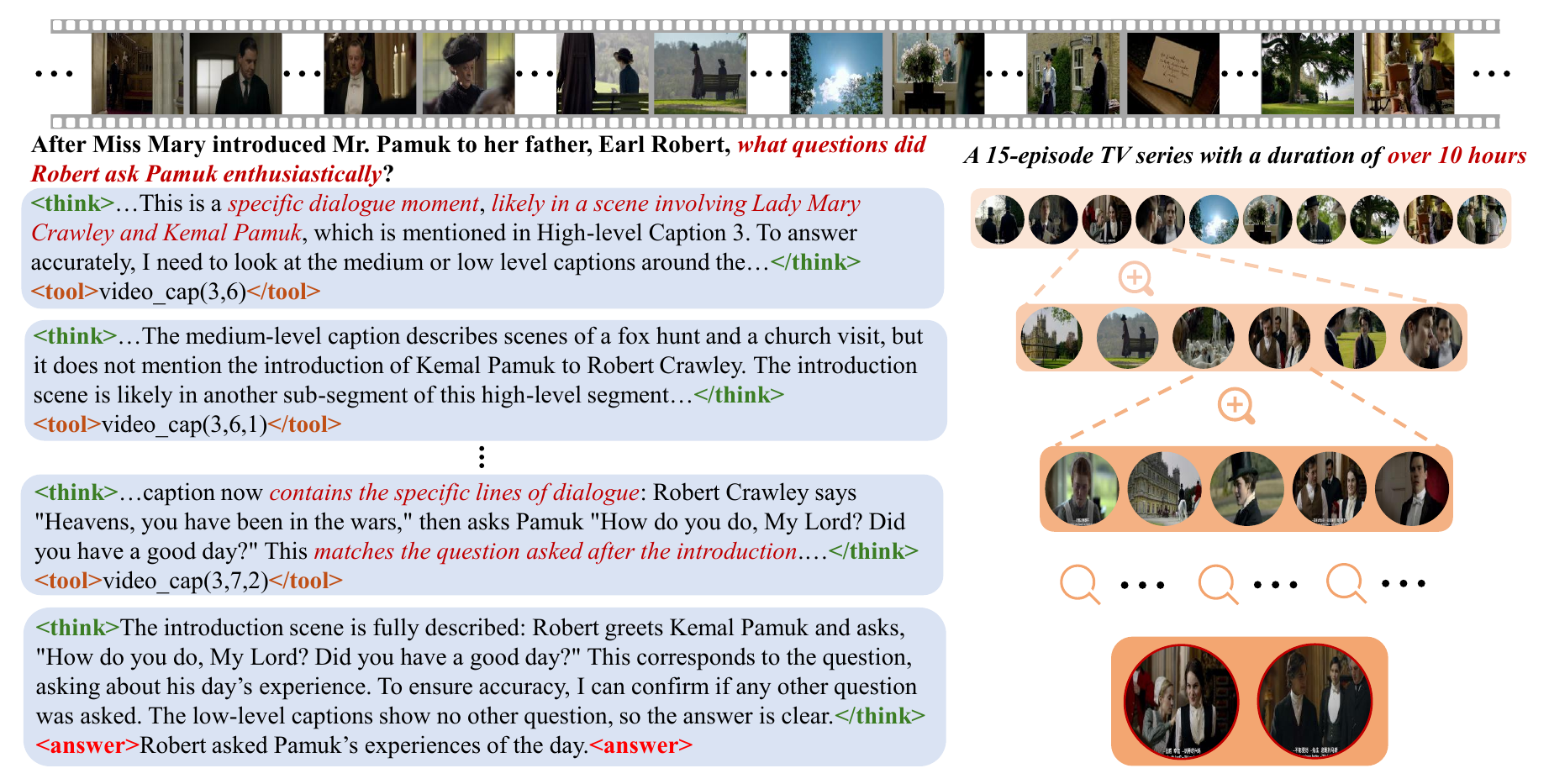}
\caption{More example on ultra-long videos.}
\label{fig:ultra_long_case_2}
\end{figure*}

\begin{figure*}
\centering
\includegraphics[width=0.99\linewidth]{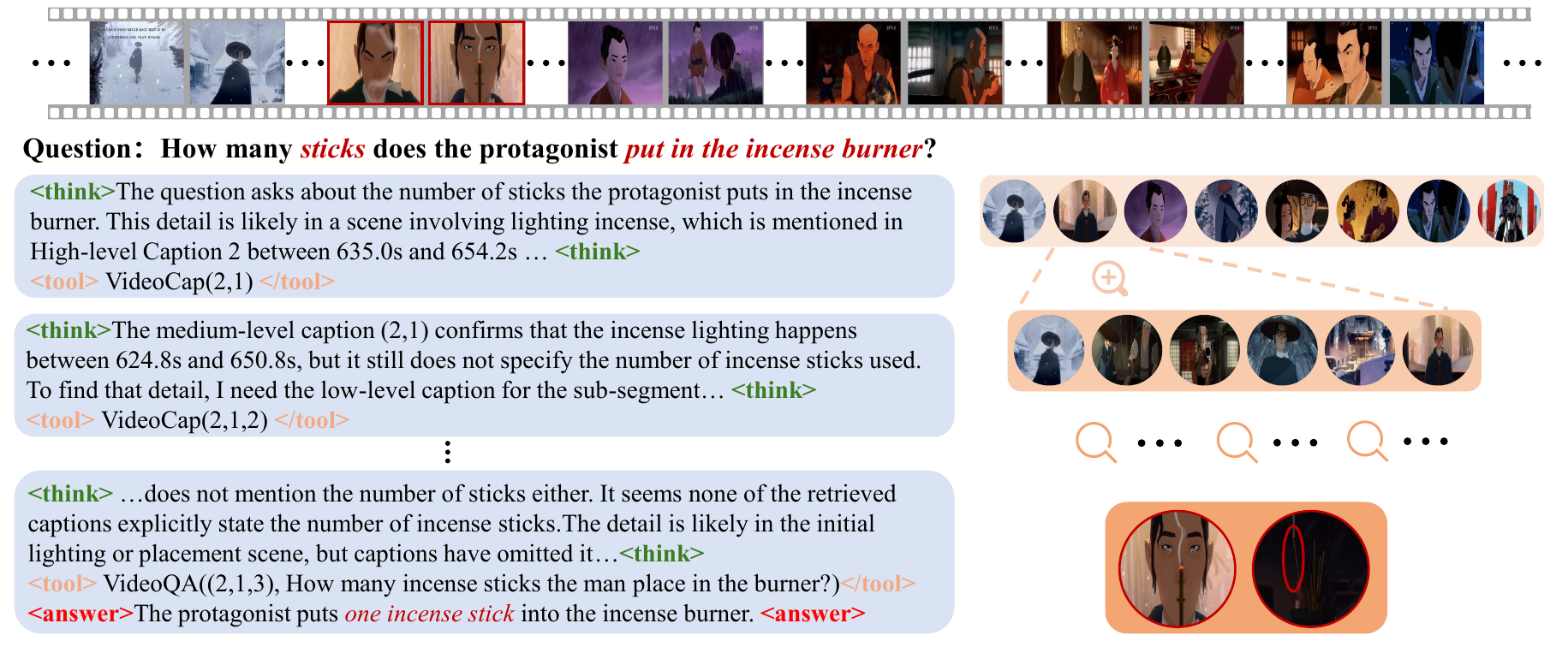}
\caption{More qualitative examples.}
\label{fig:qualitative_example_2}
\end{figure*}

\begin{figure*}
\centering
\includegraphics[width=0.99\linewidth]{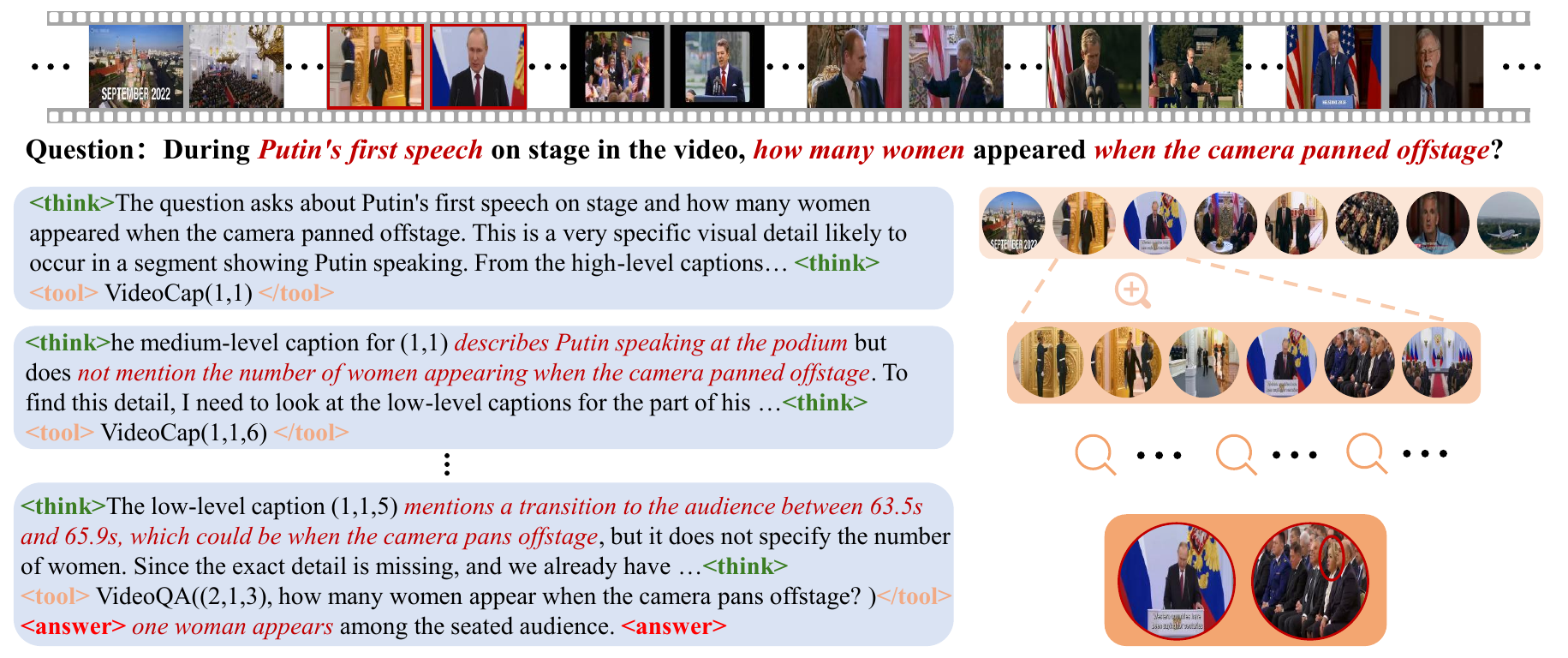}
\caption{More qualitative examples.}
\label{fig:qualitative_example_3}
\end{figure*}

We provide additional qualitative results (Figure\ref{fig:qualitative_example_2}, Figure \ref{fig:qualitative_example_3}) and ultra-long video examples (Figure \ref{fig:ultra_long_case_2}) in this section. These examples illustrate LongVideo-R1’s ability to perform hierarchical search, disambiguate similar scenes across hours-long content, and jointly use both high-level and fine-grained information.

The examples include cases from TV series such as \textit{Downton Abbey}, where the model successfully navigates multi-hour narratives, repeatedly locating the correct characters, objects, or events despite substantial visual similarity across episodes.

\section{Failure Examples}
\label{sec:appendix-failure}

\begin{figure*}
\centering
\includegraphics[width=0.99\linewidth]{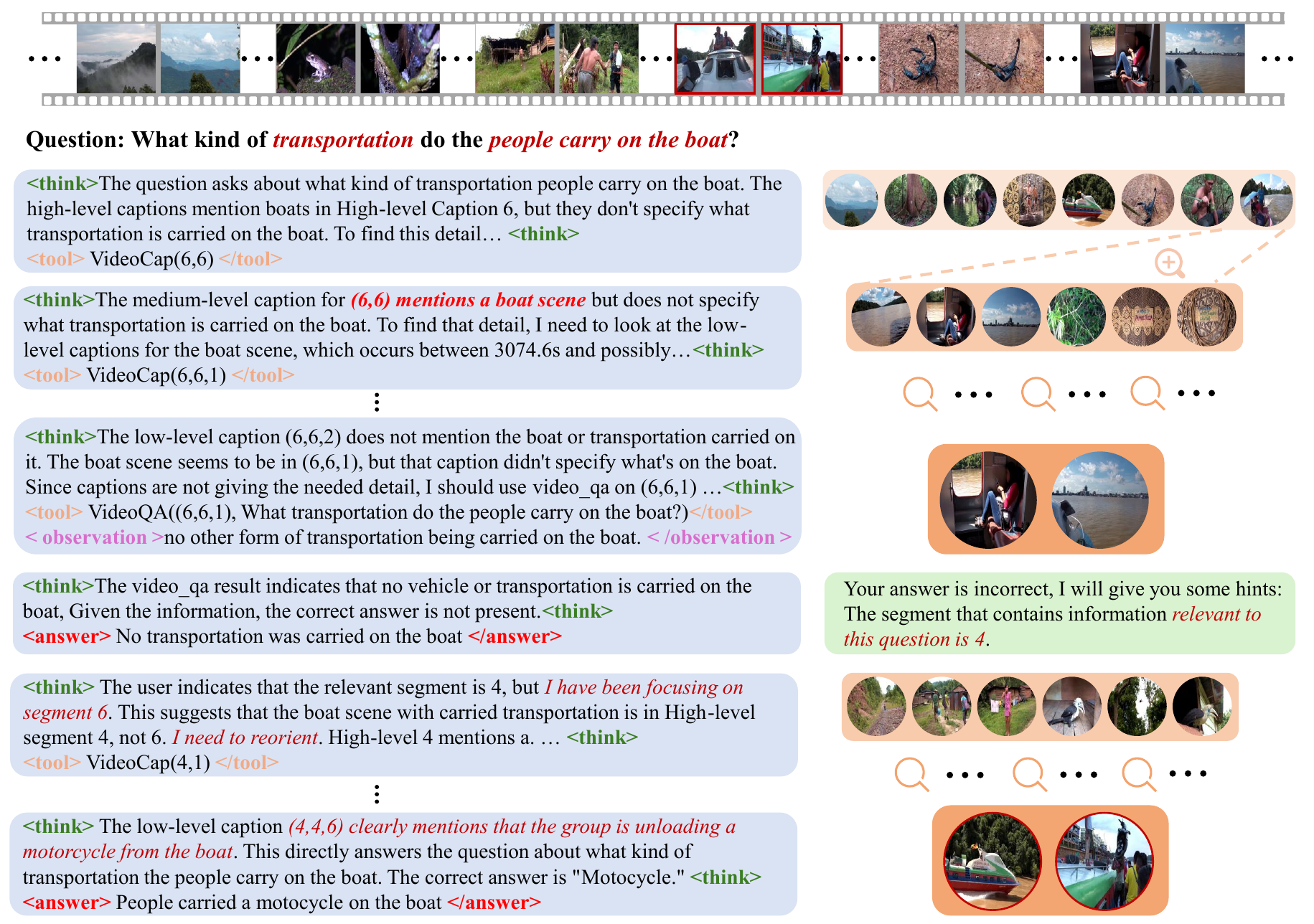}
\caption{\name may sometimes be disturbed by similar information, but people can guide the model back on track with just a few hints.}
\label{fig:failure_case_exp1}
\end{figure*}

\begin{figure*}[t]
\centering
\includegraphics[width=0.99\linewidth]{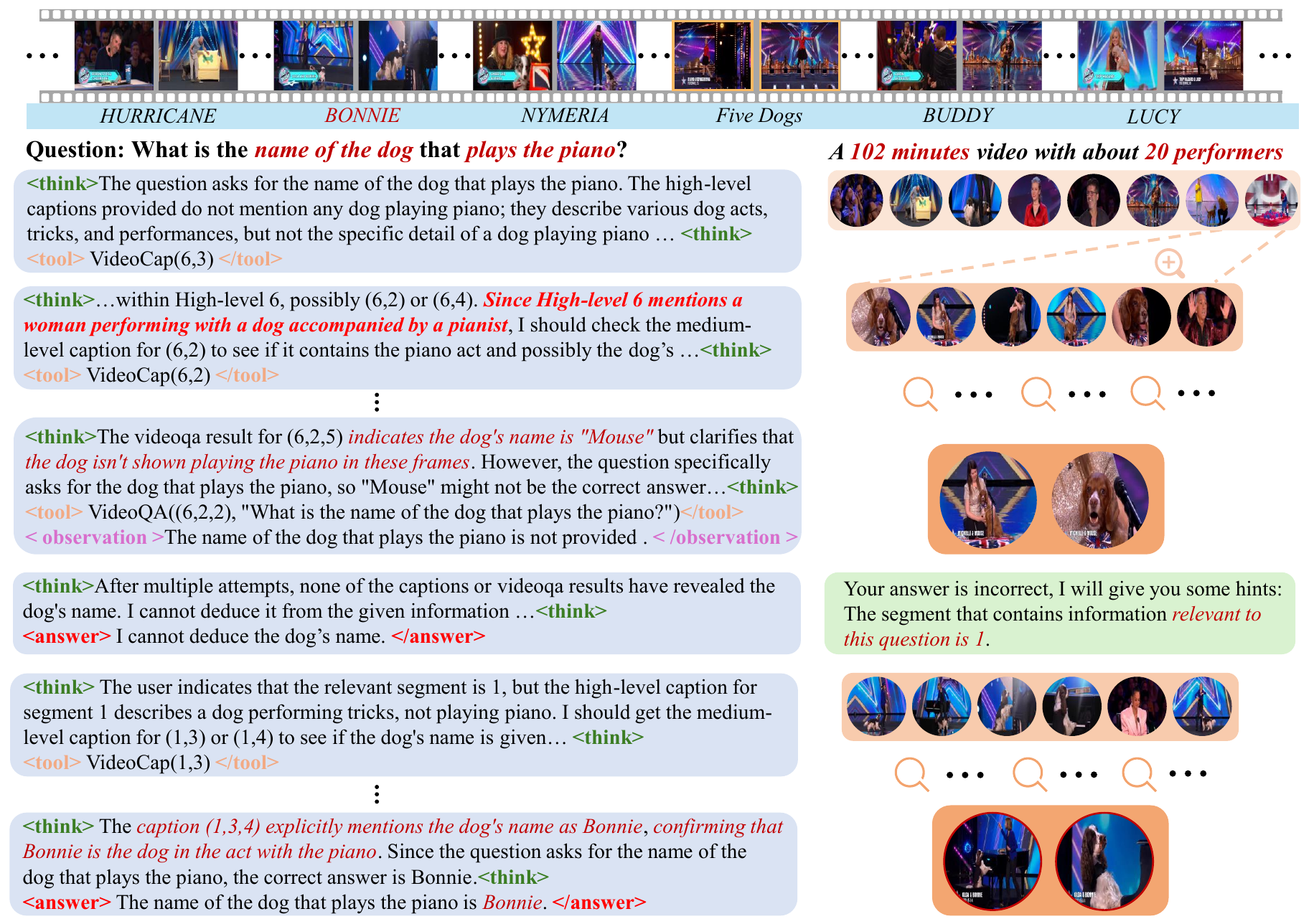}
\caption{\name may sometimes be disturbed by similar information, but people can guide the model back on track with just a few hints.}
\label{fig:failure_case_exp2}
\end{figure*}

Although LongVideo-R1 performs well across various long-video benchmarks, failure cases still occur (Figure \ref{fig:failure_case_exp1}, Figure \ref{fig:failure_case_exp2}). When a visually similar but irrelevant object appears in the video, the model sometimes commits to the wrong branch and fails to return to the correct segment.

We also find that simple textual hints can often guide the model back to the correct segment and enable it to produce the correct answer.

\end{document}